\documentclass[runningheads]{llncs}

\usepackage{eccv}

\usepackage{eccvabbrv}

\usepackage{graphicx}
\usepackage{booktabs}
\usepackage{amsmath,amssymb} 
\usepackage{xspace}
\usepackage{wrapfig}
\usepackage{multirow}
\usepackage{colortbl}
\usepackage{mathtools}
\usepackage{pifont}
\usepackage{soul}
\usepackage{amssymb}
\usepackage{xfrac}
\usepackage[title]{appendix}

\usepackage[accsupp]{axessibility}  

\usepackage{hyperref}
\usepackage[capitalize]{cleveref}

\usepackage{orcidlink}

\def\aka{a.k.a\@\xspace}
\providecommand\SHORTNAME{DAVIDE\xspace}
\providecommand\SHORTNAME{DAVIDE\xspace}
\newcommand{\xmark}{\ding{55}}%

\begin{document}

\title{DAVIDE: Depth-Aware Video Deblurring} 

\titlerunning{DAVIDE}

\author{German F. Torres\inst{1}\orcidlink{0000-0002-7555-3676} \and
Jussi Kalliola\inst{1}\and
Soumya Tripathy\inst{2} \and
Erman Acar\inst{2}\and
Joni-Kristian K\"am\"ar\"ainen\inst{1}
}

\authorrunning{G.~Torres \etal}

\institute{Tampere University, Finland \\
\email{\{german.torresvanegas, jussi.kalliola, joni.kamarainen\}@tuni.fi} \and
Huawei Technologies, Finland\\
\email{\{erman.acar, soumya.ranjan.tripathy\}@huawei.com}\\
}
\maketitle

\begin{abstract}
Video deblurring aims at recovering sharp details from a sequence of blurry frames. 
Despite the proliferation of depth sensors in mobile phones and the potential of depth information to guide deblurring, depth-aware deblurring has received only limited attention.
In this work, we introduce the 'Depth-Aware VIdeo DEblurring' (DAVIDE) dataset to study the impact of depth information in video deblurring. 
The dataset comprises synchronized blurred, sharp, and depth videos.
We investigate how the depth information should be injected into the existing deep RGB video deblurring models, and propose a strong baseline for depth-aware video deblurring.
Our findings reveal the significance of depth information in video deblurring and provide insights into the use cases where depth cues are beneficial. 
In addition, our results demonstrate that while the depth improves deblurring performance, this effect diminishes 
when models are provided with a longer temporal context. 
Project page: \url{https://germanftv.github.io/DAVIDE.github.io/}.
\end{abstract}
%
%
\section{Introduction}
\label{sec:introduction}
\begin{figure}[t]
  \centering
  \includegraphics[width=\linewidth]{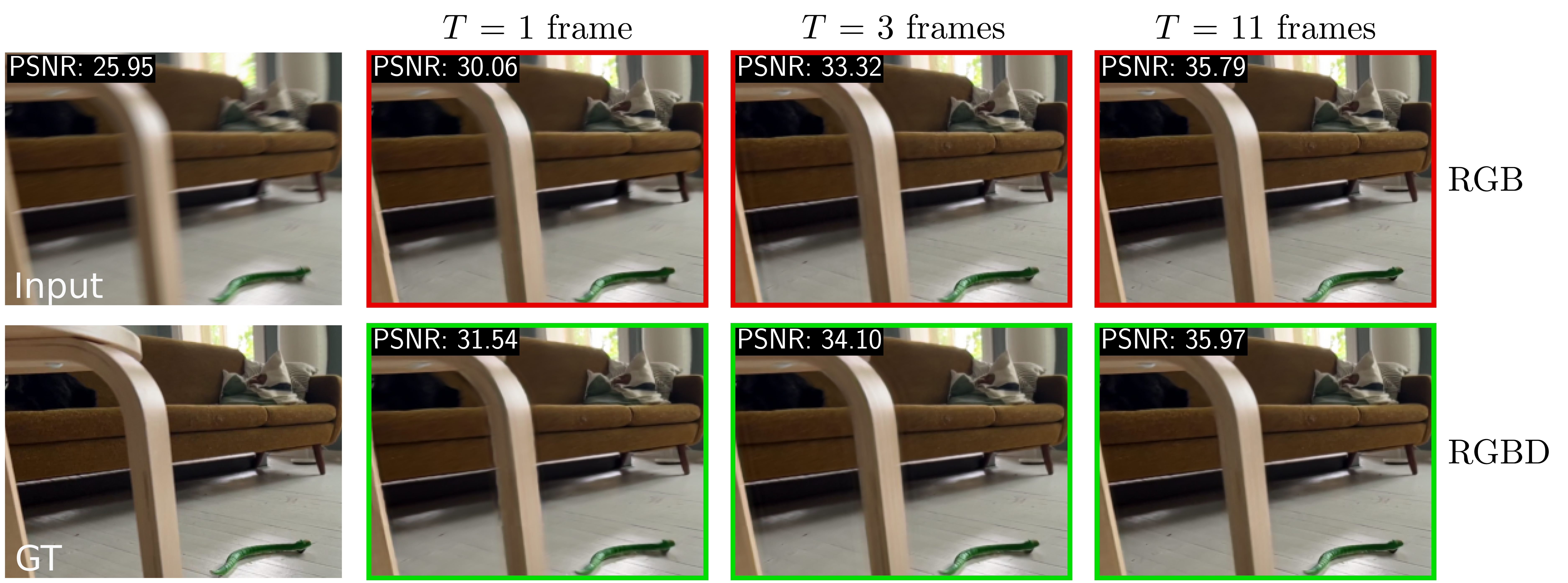}
  \caption{Examples of depth-aware video deblurring (RGBD) with increasing temporal length $T$ of the context window  (see \cref{sec:exp_depth_impact} for details).}
  \label{fig:visual_results_exp1}
\end{figure}
Motion video deblurring consists of removing the visual artifacts caused by the relative motion from the scene to the camera during the video recording.
The growing demand for slow motion and other filmographic effects justifies the development of video deblurring solutions.
State-of-the-Art deblurring methods comprise deep architectures trained in a supervised manner on large datasets containing pairs of blurry and sharp images to learn mappings from blurry to sharp.
Unlike single-image deblurring, video deblurring architectures consider temporal correlations among frames within a context window. 
This temporal information aids in reconstructing the sharp details by either implicitly or explicitly aligning and fusing data in the embedding space.

Motion blur varies spatially due to multiple factors, including depth, which is often overlooked in video deblurring research.
For instance, due to the parallax effect, objects closer to the camera exhibit more motion, and thus more blur than those farther away.
This effect is pronounced in scenarios with a moving camera and static scene, but also occurs when objects at varying depths move at the same speed, captured by a static camera; the closer objects display more motion blur.
Furthermore, depth maps indicate occlusions and depth discontinuities that should appear as sharp edges in RGB frames.
In this context, depth information could guide the deblurring process, as it provides essential cues about how the blur is formed and what is the underlying structure in the sharp image.
Accordingly, the motivation behind this work arises from the curiosity to determine the extent to which depth information can enhance performance.

Several conventional deblurring algorithms incorporate depth in their blur formation model~\cite{Xu2012, Hu2014, Park2017, Pan2019, Sheng2019}. 
Nevertheless, those only handle camera motion blur and are computationally expensive, even for single-image deblurring.
Depth has been included as an additional input in deep deblurring to enhance image quality~\cite{Li2020, Zhu2022a}.
However, these methods require initializing input depth maps by monocular depth estimation from RGB.
This raises questions about the true effectiveness of depth information when captured with a real sensor, as SotA monocular depth estimation methods do not generalize well to unseen content.
For single-image deblurring, Li~\etal~\cite{Li2020} reported improvements up to 0.64~dB in PSNR, but those results do not necessarily transfer to video deblurring as a sequence of moving camera frames provides stereo cues that deep architectures may learn to use instead of depth.
There is, therefore, a need for: 1) a large dataset with synchronized blurred, sharp, and depth videos captured by real sensors; 2) deep video deblurring methods that effectively fuse depth and RGB;
and 3) a thorough analysis of depth as an auxiliary input for video deblurring.

In this work, we address the items 1-3). We introduce a 'Depth-Aware VIdeo DEblurring' (DAVIDE) dataset for video deblurring, including synchronized blur, sharp, and depth map videos, captured with an iPhone 13 Pro that uses a LiDAR for depth sensing.
To the best of our knowledge, this is the first large-scale video debluring dataset that includes depth information, allowing training and evaluation of deep models for depth-aware video deblurring.
Secondly, we build upon a recent SotA video deblurring architecture~\cite{Li2023} and devise a depth-aware video deblurring network that processes blurry video frames and depth maps to produce sharp frames.
Specifically, we propose a depth injection method that employs the \textit{Grouped Spatial Shift} (GSS) block~\cite{Li2023} to enlarge the receptive field of depth features, along with our \textit{Depth-aware Transformer} (DaT) block for more effective integration of depth into RGB features.
Finally, we conducted a comprehensive evaluation of the role of depth in video deblurring. 
Our findings indicate that as the context window extends, video deblurring methods progressively mitigate the lack of explicit depth cues (See \cref{fig:visual_results_exp1}).
%
%
\section{Related Work}
\label{sec:related_work}
\paragraph{Video deblurring.} Video methods take advantage of the spatio-temporal correlation between consecutive frames within a 'context window' to recover the sharp details. 
Early works~\cite{Bar2007,Cho2012,Wulff2014,Hyun2015,Zhang2016} formulate deblurring as an optimization problem, 
including blur formation models and hand-crafted image priors to regularize the otherwise ill-posed problem.
These methods are computationally intensive and yield only moderate quality because of limitations in accurately modeling blur and the priors' failure to adequately represent real video characteristics.

Deep learning methods have shown superior performance in video deblurring, as they learn the mapping from blurry to sharp images from large-scale datasets.
Su~\etal~\cite{Su2017} introduce an encoder-decoder architecture that takes adjacent blurry frames and outputs their sharp estimates in an end-to-end manner. 
Due to the limited receptive field of convolution blocks, the implicit feature alignment of highly correlated but misaligned frames within the context window is challenging in the encoder-decoder architectures. 
To overcome this limitation, considerable effort has been directed towards developing effective frame alignment modules.
For example,~\cite{Kim2018, Pan2020, Xiang2020, Li2021, Lin2022, Zhang2022, Liang2022b} use optical flow to guide the alignment of the neighboring frames.
Alternatively, implicit alignment can be achieved through 3D convolution~\cite{Zhang2018, Wang2021}, 
deformable convolutions~\cite{Wang2019,Jiang2022}, or dynamic filters~\cite{Zhou2019b,Pan2023}. 
To avoid alignment, \cite{Li2021, Son2021} directly aggregate the information of a correlation volume with multiple matching candidates for each pixel.

In terms of architectural design, the
sliding window-based structure is used in many works~\cite{Su2017,Kim2018,Wang2019,Zhou2019b,Suin2021,Li2021,Zhang2022}.
Although they perform well, the structure is inefficient as each input frame is processed multiple times during inference. As a more efficient structure, \cite{Hyun2017, Zhang2018, Nah2019b, Zhong2020, Zhang2020, Son2021, Zhu2022b} adopts the recurrent structure, where information from the previous frames is propagated forward to restore the subsequent frames. 
However, recurrent methods are prone to information loss and noise amplification
due to their recurrent nature.

Recently, the Transformer architecture and its attention mechanism have been applied in video deblurring~\cite{Zhong2020,Zhang2021,Xu2021,Lin2022,Liang2022a,Liang2022b,Cao2022}.
Liang~\etal~\cite{Liang2022a} proposed a Video Restoration Transformer (VRT) that features parallel frame prediction, as opposed to sliding window-based methods.
In an effort to mitigate computational complexity, Liang~\etal~\cite{Liang2022b} incorporated a recurrent design into a transformer-based model.
However, these approaches still require large model sizes and substantial memory for processing long sequences.
Different from transformer-based designs, Li~\etal~\cite{Li2023} devised Shift-Net, a video restoration architecture based on \textit{Grouped Spatio-Temporal Shift} (GSTS).
Similarly, Pan~\etal~\cite{Pan2023} proposed a network architecture utilizing discriminative feature fusion modules and wavelet-based feature propagation.
\paragraph{Depth-Aware Deblurring.} 
The goal is to use depth as an additional cue to guide the deblurring process.
Most of the previous works are limited to \textit{single-image deblurring}.
Conventional methods adopt an alternating iterative algorithm, 
which jointly estimates the sharp image and another latent variable, 
such as the depth map or camera motion. The methods in this category 
assume a ground-truth depth map~\cite{Pan2019} or
a noisy initial depth map that is iteratively refined~\cite{Sheng2019}.
Others aim to estimate the sharp image and the depth map jointly, using a stereo setup~\cite{Xu2012}, 
an image sequence~\cite{Park2017},
or exploiting the underlying geometrical relationships between the clear image and the depth of the scene that produce motion blur~\cite{Hu2014}.
In terms of visual quality, these methods produce only moderate results,
as they rely on traditional deconvolution techniques~\cite{Scales1988,Chambolle2011,Krishnan2009}.
In~\cite{Torres2023} the dept-aware camera motion blur is modelled more precisely but their method
assumes that the camera motion trajectory is available.

As a deep learning architecture,
Li~\etal~\cite{Li2020} proposed a deblurring network that takes as input the depth map and the blurry image, and outputs the sharp image. 
Their network performs favorably in single-image deblurring, 
but does not scale well to video deblurring since it only concatenates consecutive frames into a 3D tensor for network input.
Inspired by the EDVR~\cite{Wang2019} architecture, 
Zhu~\etal~\cite{Zhu2022a} devised a depth-aware video deblurring neural network that outperforms methods that do not incorporate depth.
Notably, both above architectures necessitate initializing depth maps through a monocular depth estimation method from RGB frames and do not experiment on real depth maps,
leaving the true impact of depth captured by a dedicated depth sensor uncertain.
Feng~\etal~\cite{Feng2023} proposed a video enhancement network integrating sparse depth and IMU information to improve the quality of the degraded video, 
The KITTI dataset~\cite{Geiger2013} is considered in their experiments, as it incorporates LiDAR depth and IMU data.
However, the blur is synthetic and fails to represent realistic motion blur.
\paragraph{Deblurring benchmark datasets.} To the authors' best knowledge, no public datasets for depth-aware video deblurring exist.
In fact, assembling a dataset for video deblurring even without depth information is already a challenging task since two cameras and an optical beam splitter are needed.
RealBlur~\cite{Rim2020} and BSD~\cite{Zhong2023} are datasets that feature real recorded blur with ground-truth.
Many datasets circumvent the complex two-camera setup by averaging frames from high frame-rate videos.
The most popular benchmark datasets are GoPro~\cite{Nah2017}, DVD~\cite{Su2017}, 
REDS~\cite{Nah2019a}, and HIDE~\cite{Shen2019}. None of the above datasets includes depth.

In principle, video averaging could be applied to RGB-D video datasets such as TUM RGB-D~\cite{Sturm2012}, NYU Depth~\cite{Silberman2012}, Cityscapes~\cite{Cordts2016}, or KITTI~\cite{Geiger2013} to derive sharp, blur, and depth frames. 
However, these datasets are either too small, offer limited scene variability (\eg, self-driving scenarios), or have a too low frame rate. 
Due to these limitations, the DAVIDE dataset is introduced in this work.
%
%
\section{Depth-Aware VIdeo DEblurring dataset (DAVIDE)}
\label{sec:dataset}
\SHORTNAME follows the construction steps of the REDS dataset for video deblurring~\cite{Nah2019a}: 1) data recording, 2) frame interpolation, 3) camera response calibration, 4) blur synthesis, and 5) splitting to train, validation, and test sets.

\subsection{Background}
Dynamic blur can be synthesized by averaging high rate video frames~\cite{Nah2017,Su2017,Nah2019a,Shen2019}.
The original frames $I[k]$, $k=0,1,2,\ldots$ are averaged to produce
blurry frames $I_b[m]$, $m=0,1,2,\ldots$
\begin{equation}
    I_b[m] = \textit{CRF}\left(\frac{1}{N}\sum_{n=0}^{N-1} I_L[mN+n] \right) \quad .
\label{eq:averaging}
\end{equation}
In \cref{eq:averaging}, $N$ is the number of averaged frames, 
$\textit{CRF}$ is a non-linear Camera Response Function, and $I_L$ are sharp high-speed
camera frames in the linear color space obtained by inverse CRF, $I_L[k]=\textit{CRF}^{-1}\left(I[k]\right)$ (\cref{sec:blur_synthesis}). 
For each synthesized blurry frame $I_b[m]$, the middle original frame is the sharp groundtruth
\begin{equation}
    I_s[m] = I[mN + \lfloor N/2 \rfloor]
\label{eq:middle_frame}
\end{equation}

\subsection{Data recording}
Any RGBD sensor that provides registered and synchronized RGB frames and depth maps is suitable for data capture, as long as it has a sufficiently high frame rate and RGB quality. We selected a high-end mobile phone that captures high-quality RGB frames at 60~fps. In addition, the phone allows storage of depth maps produced by its depth estimation pipeline. The pipeline combines real depth measurements, via an on-device LiDAR sensor, and monocular depth estimation.
We implemented an iOS app for data capture and deployed it on an iPhone 13 Pro. 
The iOS app captures and stores aligned and synchronized RGB frames and depth maps at 60~fps. The RGB resolution is the sensor's native 1920x1440 pixels, and the depth maps are 256x192. 
Additionally, the App stores confidence maps, camera poses, and IMU measurements.
The confidence maps, sourced from the ARKit library, provide reliability values of the LiDAR depth measurements, particularly less accurate on highly reflective or absorbent surfaces.
These maps, along with camera poses and IMU data from ARKit and CoreMotion libraries, are included in the DAVIDE dataset to encourage further research, although we do not utilize the pose and IMU data in this work.

The captured videos represent natural indoor and outdoor camera movements when tracking various moving targets.
No identifiable information, such as people's faces or other recognizable identifiers,
was included to comply with the GDPR.
Since the frame rate was hardware-limited to 60~fps, we carefully selected clips without too fast motion. Later, all sequences were manually checked and videos with notable motion blur were removed.
\begin{figure}[t]
  \centering
  \includegraphics[width=0.9\linewidth]{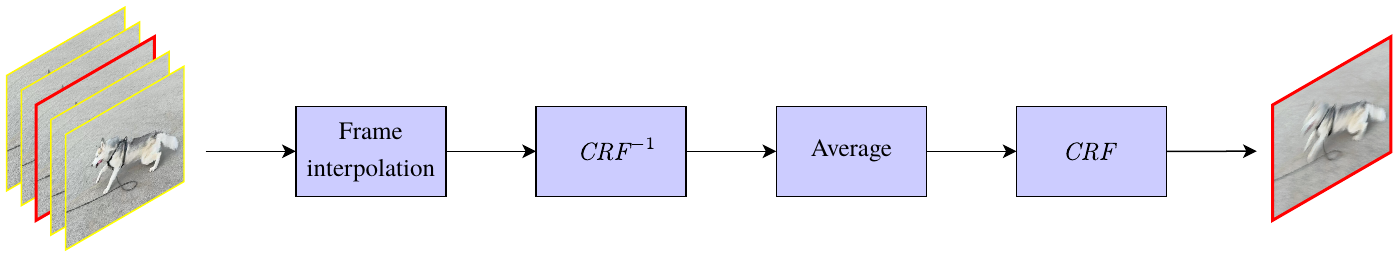}
  \caption{The \SHORTNAME blur synthesis pipeline.}
  \label{fig:dataset_synthesis_pipeline}
\end{figure}
\subsection{Blur synthesis}
\label{sec:blur_synthesis}
The blur synthesis pipeline of \SHORTNAME is depicted in \cref{fig:dataset_synthesis_pipeline}. The first step after data recording is \textit{frame interpolation}. Frame interpolation is needed to produce more natural and smooth blur, since otherwise the averaging in \cref{eq:averaging} can produce 'ghost images' of fast moving objects~\cite{Nah2019a}. The deep learning-based frame interpolation methods VFI-ASC\cite{Niklaus-2017-iccv} (used in REDS~\cite{Nah2019a}), SS-SloMo~\cite{SSloMo}, XVFI~\cite{Sim2021}, and EMA-VFI~\cite{EMA-VFI} were tested. XVFI was found the best and then used to
generate 7 intermediate frames between each two original frames.
This procedure increased the time resolution 8 times corresponding to 480~fps.

\textit{Camera Response Function (CRF)} and its inverse are needed to map the pixel colors to a linear color space for \cref{eq:averaging}.
Standard Gamma function and its inverse were used in the GoPro dataset~\cite{Nah2017}, but the CRF can be calibrated for a known sensor by applying the Robertson's~\cite{Robertson-CRF} or Debevec's~\cite{Debevec-CRF} methods. Robertson's was used in REDS~\cite{Nah2019a}, but we found Debevec more straightforward and as it produces strictly monotonic mapping.
The details of the CRF calibration process are described in \cref{sec:crf_calibration}.
\textit{Blur synthesis} was performed according to \cref{eq:averaging} by averaging interpolated 480 fps video to create a blurry virtual video of 15 fps. 
Sharp and depth correspondences were derived from the middle frame, as specified in \cref{eq:middle_frame}.
\subsection{Dataset details}
\label{sec:dataset_details}
\begin{table}[b]
\caption{Details of the \SHORTNAME test clips and the 7 annotated attributes (CM: camera motion; MO: moving objects).}
\label{tab:test_split_attributes}
\centering
\resizebox{0.72\linewidth}{!}{
\begin{tabular}{c cc c cc c ccc}
\toprule
\multirow{2}{*}{Test set} & \multicolumn{2}{c}{\textit{Environment}} & \enspace & \multicolumn{2}{c}{\textit{Motion}} & \enspace & \multicolumn{3}{c}{\textit{Proximity}} \\
 & Indoors & Outdoors & & CM & CM+MO & & Close & Mid & Far \\
\midrule
\# of clips & 4 & 10 & & 4 & 10 & & - & - & - \\
\# of frames & 1,043 & 2,627 & & 1,249 & 2,421 & & 1,363 & 1,481 & 826 \\
\bottomrule
\end{tabular}}
\end{table}
Original videos containing blurry frames and interpolated frames with an excessive amount of artifacts were removed. 
In the end, the final DAVIDE dataset comprises 90 clips divided into 69 (16,106 frames) for training, 7 (1,669 frames) for validation, and 14 (3,670 frames) for testing.
In the test split (14 clips), each frame was annotated with seven content attributes (see~\cref{tab:test_split_attributes}), categorized by: 1) environment (indoors/outdoors), 2) motion (camera motion/camera and object motion), and 3) scene proximity (close/mid/far).
The 'environment' and 'motion' categories were determined manually for each clip, and 'proximity' values were determined using an automated procedure for each frame. The procedure involves segmenting the depth map into three distance bins:
$(0-1.5]$ (Close), $(1.5-4.5]$ (Mid), and $(4.5-$ (Far) meters. Each depth map pixel was assigned to one of the three bins, and the largest bin was used to assign the attribute value.
These annotations aim to facilitate further analysis into scenarios where depth information proves most beneficial.
%
%
\section{Method}
\label{sect:method}
Shift-Net~\cite{Li2023} was selected as the base model for our depth-aware video deblurring. 
It performs well in multiple video restoration tasks and is more compact than the competing ones~\cite{Liang2022a, Zhu2022b, Chen2022, Wang2019}.
Shift-Net utilizes the \textit{Grouped Spatial-Temporal Shift} (GSTS) block to implicitly aggregate correspondences among the neighboring input frames.
This block, with minimal computation, provides a wide receptive field for efficient multi-frame fusion and has been proven effective.
\begin{figure}[b]
    \begin{center}
    \includegraphics[width=0.98\linewidth]{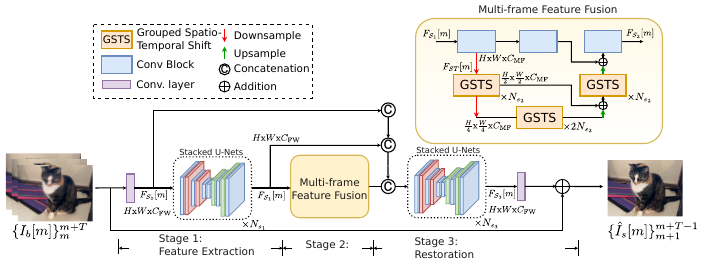}
    \end{center}
    \caption{Overview of Shift-Net.}
   \label{fig:ShiftNet_overview}
\end{figure}
\subsection{Overview of Shift-Net}
\label{sec:shift-net_overview}
Given a blurry sequence $\{I_b[m] \in \mathbb{R}^{H \times W \times 3}\}_m^{m+T}$, where $H \times W$ denotes the image resolution and $T$ is the temporal length of the context window, Shift-Net produces a sequence of sharp estimates $\{\hat{I}_s[m] \in \mathbb{R}^{H \times W \times 3}\}_{m+1}^{m+T-1}$.
Shift-Net operates in three stages (\cref{fig:ShiftNet_overview}): 1) feature extraction, 2) multi-frame feature fusion, and 3) restoration.
The feature extraction and restoration are performed by stacks of $N_{s_1}$ and $N_{s_3}$ 3-level U-Nets, respectively. The output of stage 1 is a feature tensor with $C_\textbf{FW}$ channels for each frame. Stage 2 has a stack of $N_{s_2}$ \textit{Grouped Spatial-Temporal Shift} (GSTS) blocks that establish temporal feature correspondences at different spatial resolutions with $C_{\textbf{MF}}$ feature channels.
\paragraph{Grouped Spatial-Temporal Shift (GSTS).}
A GSTS block consists of a grouped spatio-temporal shift operation followed by a lightweight fusion layer (based on the NAFNet~\cite{Chen2022}). 
While the shift operation mixes the features across adjacent frames and channels, the fusion layer aggregates the information from these mixed features.
A single spatial-temporal shift can process only two adjacent frames, and
therefore GSTS blocks alternate forward and backward spatio-temporal shifts to establish bidirectional aggregation.
We outline the spatio-temporal shift algorithmically; for further details, see the original paper~\cite{Li2023}.

The feature sequence $\{F_{ST}[m]\}_m^{m+T}$ in GSTS is reshaped to a 4D tensor $F_{ST} \in \mathbb{R}^{T \times C_{\text{MF}} \times \hat{H} \times \hat{W} }$, where $\hat{H} \times \hat{W}$ is the level resolution.
The tensor is temporally shifted by $\pm \frac{C_{\textbf{MF}}}{2}$,
\begin{align}
\label{eq:temporal_shift}
\begin{split}
    F_{ST} &\rightarrow \mathbb{R}^{(T \cdot C_{\textbf{MF}}) \times \hat{H} \times \hat{W}}, \quad \text{\texttt{\# reshape tensor}} \\
    F_{ST} &\coloneq \texttt{roll}(F_{ST}, \pm \sfrac{C_{\textbf{MF}}}{2}, \text{dim}=1) \quad \text{\texttt{\# apply temporal shift}}\\
    F_{ST} &\rightarrow \mathbb{R}^{T \times C_{\textbf{MF}} \times \hat{H} \times \hat{W}}, \quad \text{\texttt{\# reshape back}} 
\end{split}
\end{align}
where "$+$" and "$-$" denote the forward and backward shift direction.
\cref{eq:temporal_shift} effectively shifts each channel of the frame halfway across the total number of channels, mixing temporal information across adjacent frames and channels.

\noindent\textbf{Grouped Spatial Shift (GSS).} Following that, a subset of channels $F_{ss} \in F_{ST}$ is selected for spatial shifting, either the first half or the last half, depending on the forward/backward direction.
The channels are spatially shifted by $(\Delta x, \Delta y)$, specific for each channel group:
\begin{equation}
\label{eq:spatial_shift}
    F'_{ss} = \texttt{GroupedSpatialShift}(F_{ss}, \Delta x, \Delta y)
\end{equation}
Prior to the lightweight fusion, the features $F_{ss}$ and $F'_{ss}$ are concatenated.
\subsection{Depth injection}
\label{sec:depth_injection}
\begin{wrapfigure}{l}{0.25\textwidth}
  \begin{center}
   \vspace{-5\medskipamount}
    \includegraphics[width=\linewidth]{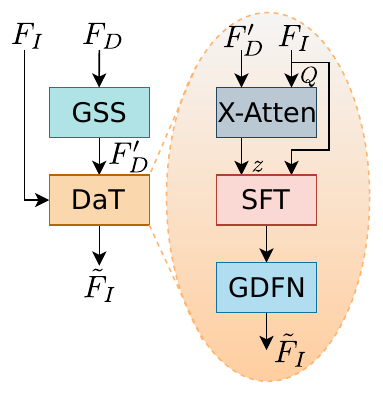}
  \end{center}
  \caption{\small Depth fusion block.}
  \label{fig:depth_fusion}
\end{wrapfigure}%
We first replicate the Shift-Net's RGB processing blocks of stages 1 and 2 to extract features for the depth itself. Then, depth fusion blocks are applied at several points of the RGB features to integrate the relevant information of the depth.
Specifically, we add these depth fusion blocks after shallow and deep feature extraction in stage 1, and at each level of the stage 2 decoder multi-frame fusion.

The two previous works~\cite{Li2020,Zhu2022a} on depth-aware image/video deblurring both utilize the Spatial Feature Transform (SFT) layer~\cite{Wang2018} for depth fusion. 
We propose another extension that incorporates GSS and our \textit{Depth-aware Transformer} (DaT) block.
GSS expands the receptive field of depth features with spatial shift, while DaT more effectively aggregates features to capture depth cues.
\paragraph{Depth-aware Transformer Block (DaT).}
The DaT structure (\cref{fig:depth_fusion}) is inspired by the Restormer architecture~\cite{Zamir2022} and the SFT layer.
The fusion principle in SFT is to modulate the RGB features with an affine function of scale $\gamma$ and offset $\beta$ that are predicted through convolution layers conditioned by the depth features $z$. 
Our DaT block adapts the conditioned features with a cross-attention module ('X-Atten' block in \cref{fig:depth_fusion}) and performs feature aggregation with a gated feed-forward network~\cite{Zamir2022} ('GDFN' block in \cref{fig:depth_fusion}).
The exact details of the DaT block are in \cref{sec:dat}.
%
%
\section{Experimental results}
\label{sec:experiments}
\noindent\textbf{Implementation details.} 
We trained both RGB-only Shift-Net and our RGBD extension with stack sizes $N_{s_1} = 2$, $N_{s_2} = 2$, and $N_{s_3} = 2$; and channel dimensions $C_\textbf{FW}=16$ and $C_\textbf{MF}=64$.
For data augmentation, we used horizontal and vertical flips, using a patch size of $256 \times 256$.
The models were trained for 200 epochs with a batch size of 4 using an AdamW optimizer.
The learning rate was reduced from $3 \times 10^{-3}$ to $1 \times 10^{-7}$, using a cosine annealing strategy.
For evaluation, we adopted the standard Peak-Signal-to-Noise-Ratio (PSNR) and Structural Similarity (SSIM) index used in the previous benchmark datasets. 
\subsection{Impact of the depth cue}
\label{sec:exp_depth_impact}
\begin{figure}[b]
    \begin{tabular}{cc}
    \begin{minipage}[c]{.34\linewidth}
        \centering
        \includegraphics[width=\linewidth]{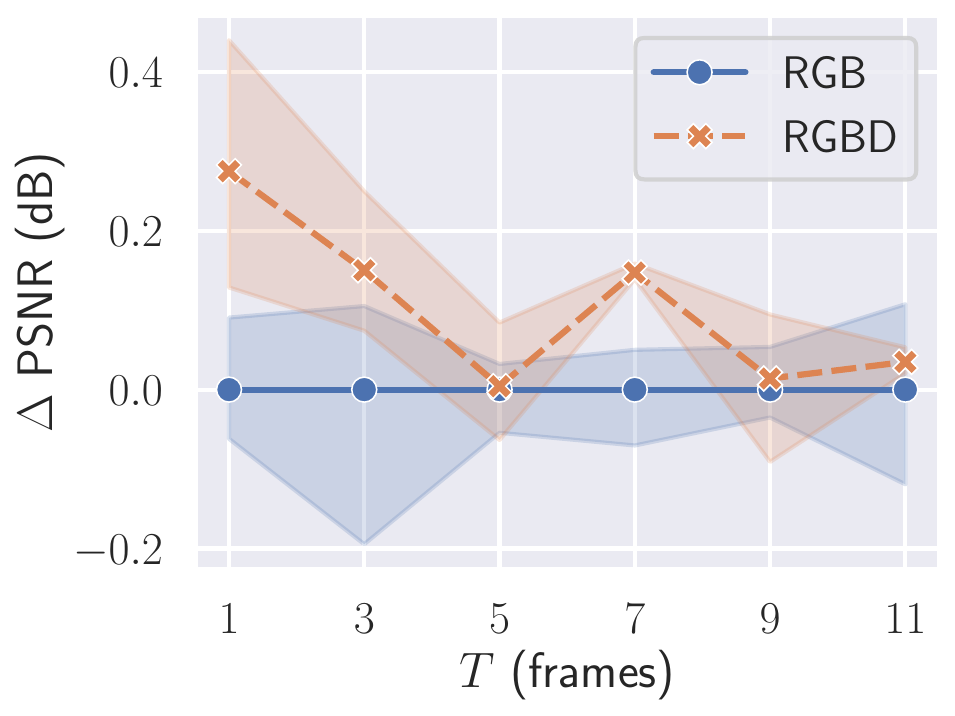}
    \end{minipage} &
    \begin{minipage}[c]{.6\linewidth}
    {\scriptsize Average PSNRs (dB) of Shift-Net models for the DAVIDE test set.}
        \centering
        \resizebox{0.95\linewidth}{!}{
        \begin{tabular}{lcccccc}
            \toprule
            $T$ (frames) & 1 & 3 & 5 & 7 & 9 & 11 \\
            \midrule
            RGB & 25.279 & 27.318 & 28.837 & 28.978 & 29.207 & 29.293 \\
            RGBD & 25.554 & 27.468 & 28.841 & 29.126 & 29.221 & 29.328 \\
            \midrule
            Diff. ($\pm$dB)     & +0.275 & +0.150 & +0.004 & +0.148 & +0.014 & +0.035 \\
            \bottomrule
        \end{tabular}}
    \end{minipage}\\
      (a) & (b) \\
   \end{tabular}
    \caption{Impact of the depth cue with varying $T$ (temporal length of the context window); (a) 95\% confidence ($\pm$2 std) plot between the RGB and RGBD performance; (b) avg. results over three independent runs.}
    \label{fig:exp_multiframe}
\end{figure}
While depth informs about sharp edges, learning-based methods may extract the same information from multiple frames.
In this experiment, we investigated the relative impact of adding depth information versus extending the context window in the input sequence.
To study this, we trained the original RGB Shift-Net and our RGBD variant three times each, with varying temporal lengths $T$ of context window in the blurry input sequence (\cref{sec:shift-net_overview}).
The performance numbers in \cref{fig:exp_multiframe}(b) show that depth contributes to deblurring performance, but the contribution quickly diminishes when the context window has more than 5 frames. The gain was +0.275 dB for $T=1$ and +0.150 dB for $T=3$ with clear margins (see~\cref{fig:exp_multiframe}(a)), but negligible for $T\geq 5$ frames ($T=7$ being a positive outlier). Visual examples illustrating this behavior are shown in  \cref{fig:visual_results_exp1}.
For $T=1$, our RGBD Shift-Net model can better recover the sharp details of the front chair and the sofa in the background compared to its RGB-only counterpart, probably leveraging the geometric information available from the depth. For $T=3$, while still beneficial, the advantage of depth information diminishes. Meanwhile, for $T=11$, the results of RGBD and RGB-only are perceptually indistinguishable. 
This shows that Shift-Net can compensate for the absence of explicit depth cues as the temporal length of the context window $T$ increases.
\paragraph{Depth reliability.}
\begin{figure}[t]
     \centering
     \begin{subfigure}[b]{0.3\textwidth}
         \centering
         \includegraphics[height=1.05\textwidth]{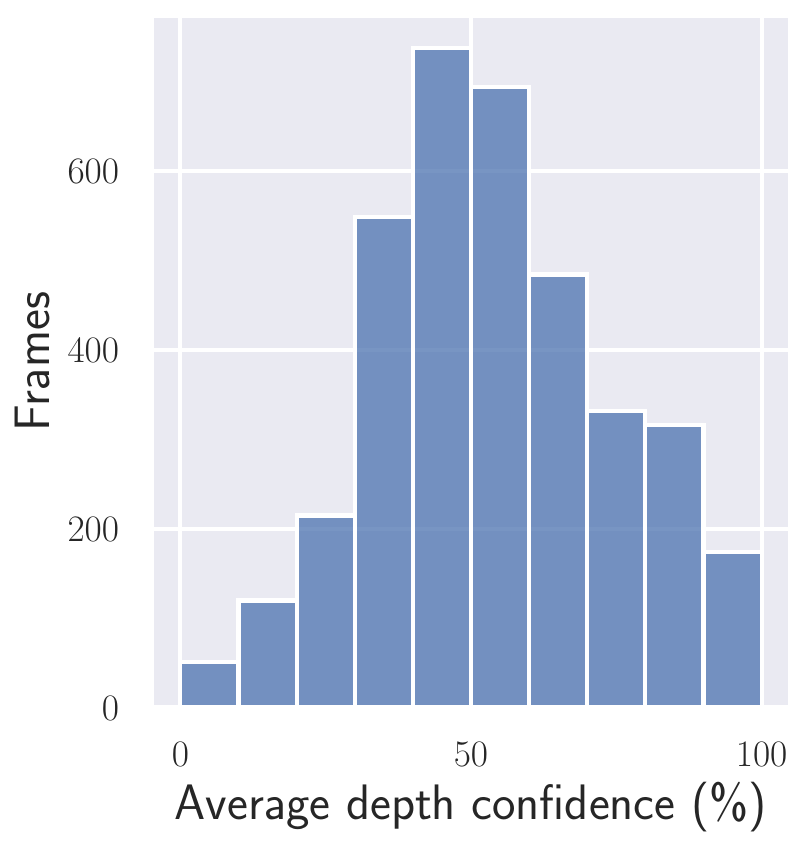}
         \caption{ }
     \end{subfigure}
     \hfill
     \begin{subfigure}[b]{0.59\textwidth}
         \centering
         \includegraphics[height=0.55\textwidth]{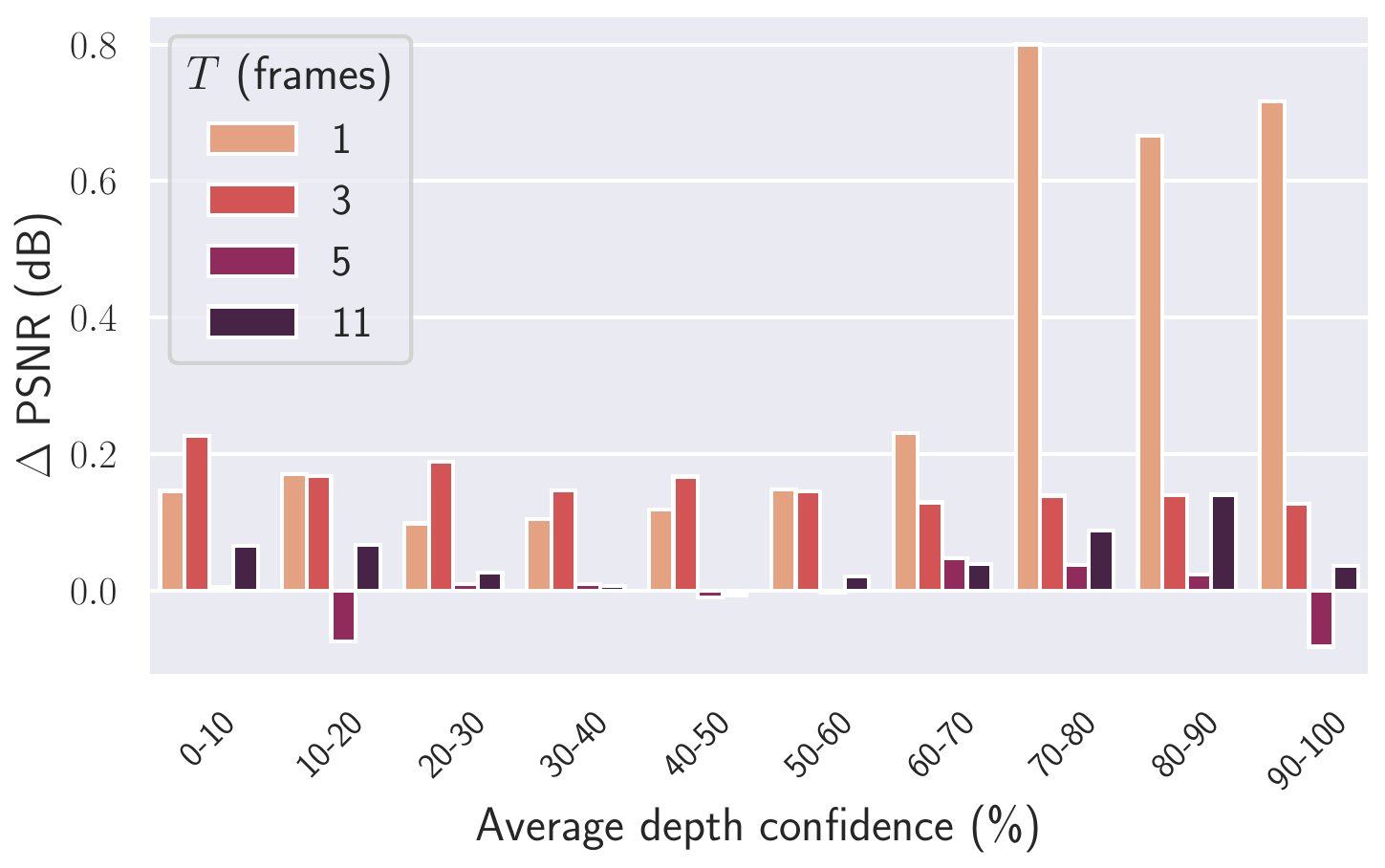}
         \caption{ }
     \end{subfigure}
        \caption{Effect of unreliable depth across confidence bins. (a) Histogram of average depth frame confidence in the test set. (b) PSNR gain across confidence ranges}
        \label{fig:depth_reliability}
\end{figure}
We utilized the confidence maps provided in the DAVIDE dataset to analyze the effect of feeding unreliable depth measurements to our extended RGBD Shift-Net.
\cref{fig:depth_reliability}(a) shows the histogram of the average confidence in the test set, which roughly follows a Gaussian distribution centered around 50\%, with a slightly higher proportion of frames exceeding 50\% confidence. 
Then we aggregated the PSNR gains in various confidence bins, as shown in
\cref{fig:depth_reliability}(b). For $T \geq 5$, where depth's contribution to Shift-Net is minimal, the gains are small or sometimes negative, exhibiting no clear pattern. 
However, for $T=3$, all gains are positive and evenly distributed. 
Particularly, for $T = 1$, depth information significantly improves deblurring at confidence levels of 70\% or higher, with no negative impact for lower confidence levels.
\paragraph{Dataset attributes.} 
\begin{figure}[t]
     \centering
     \begin{subfigure}[b]{0.3\textwidth}
         \centering
         \includegraphics[height=0.84\textwidth]{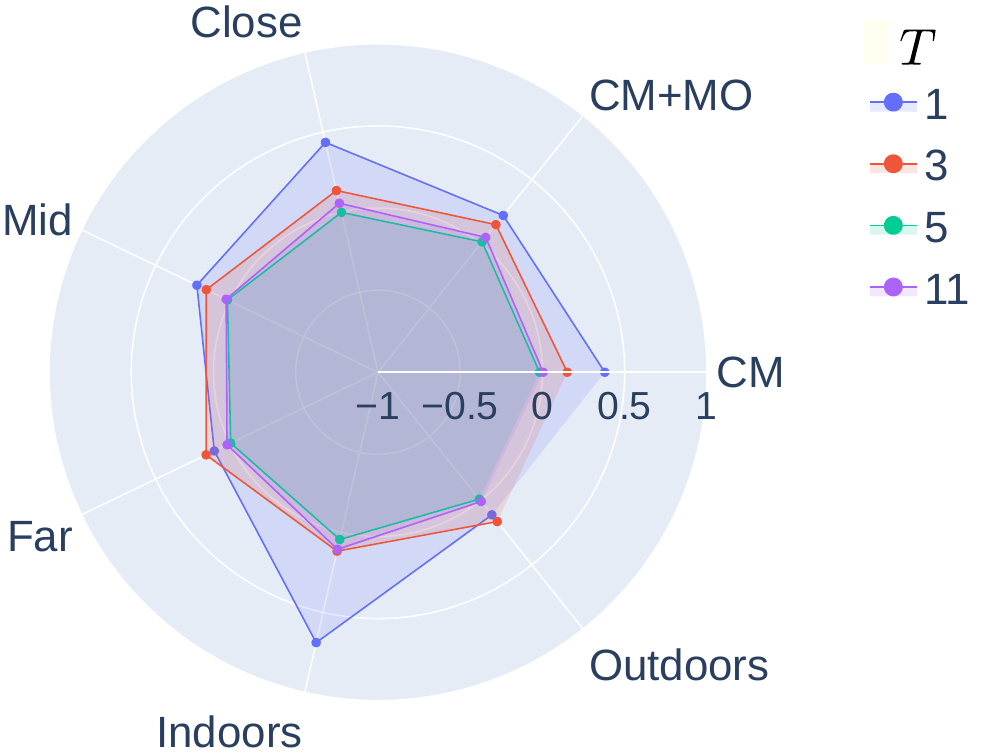}
         \caption{ }
     \end{subfigure}
     \hfill
     \begin{subfigure}[b]{0.6\textwidth}
         \centering
         \includegraphics[height=0.48\textwidth]{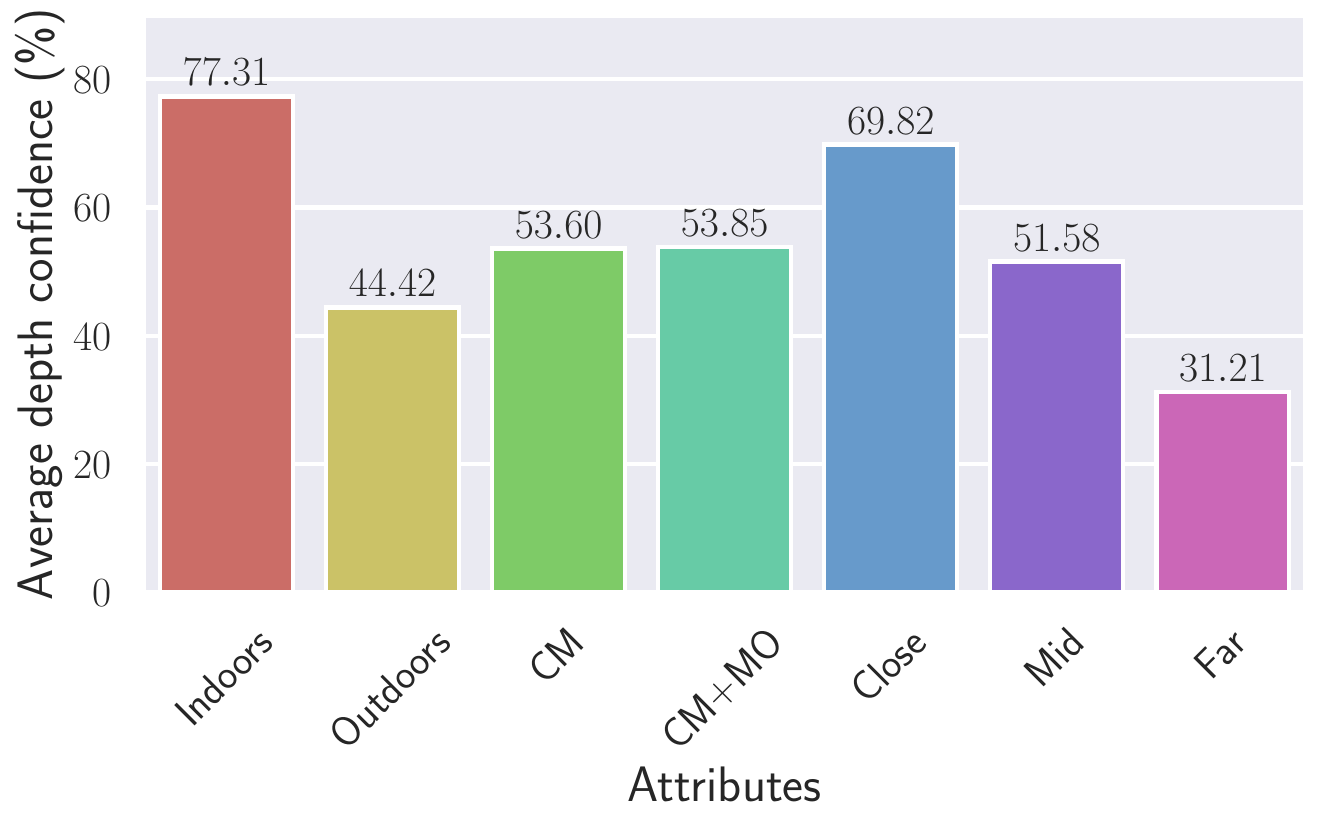}
         \caption{ }
     \end{subfigure}
        \caption{Impact of the depth cue across attributes (see \cref{tab:test_split_attributes}). (a) PSNR gains. (b) Average confidence depth per attribute.}
        \label{fig:evaluation_by_attributes}
\end{figure}
The DAVIDE test set was annotated with attributes in the hope of finding the cases where depth is most helpful (\cref{sec:dataset_details}).
\cref{fig:evaluation_by_attributes}(a) illustrates the PSNR gains for adding our depth block to Shift-Net across the annotated attributes. 
As already observed from the average results, the depth contribution diminishes when the context window has more than 5 frames. 
For $T=3$, the gains are uniform across the attributes, but the single-frame case ($T=1$) shows interesting differences. 
The three cases that benefit from depth are i) indoor, ii) close proximity, and camera motion (CM) scenes, with indoors standing out the most.
This could justified by the fact that depth sensors are more accurate at close-range distances and in indoor scenes, as shown in \cref{fig:evaluation_by_attributes}(b). 
Refer to the video samples in the project page for visual examples.
\subsection{State-of-the-Art comparison}
\label{sec:exp_sota}
We evaluated the original RGB Shift-Net and our RGBD extension against various SotA single-image and video deblurring models.
We included RGB-only video-based methods such as EDVR~\cite{Wang2019}, RVRT~\cite{Liang2022b}, and VRT~\cite{Liang2022a}.
For depth-aware deblurring, we only considered the DGN~\cite{Li2020}, which we trained for both single-image and video deblurring, using $T=1$ and $T=5$, respectively (values taken from the original paper~\cite{Li2020}). 
We excluded the method from~\cite{Zhu2022a} due to its inaccessible implementation.
All models were trained from scratch. 
The training parameters were taken from the original publications, including the temporal length $T$ of the context window.
Only a few adjustments to the original settings were required to ensure solid convergence.
Detailed training settings for these models are provided in \cref{sec:sota_training_settings}.

\begin{figure}[t]
    \begin{center}\includegraphics[width=\linewidth]{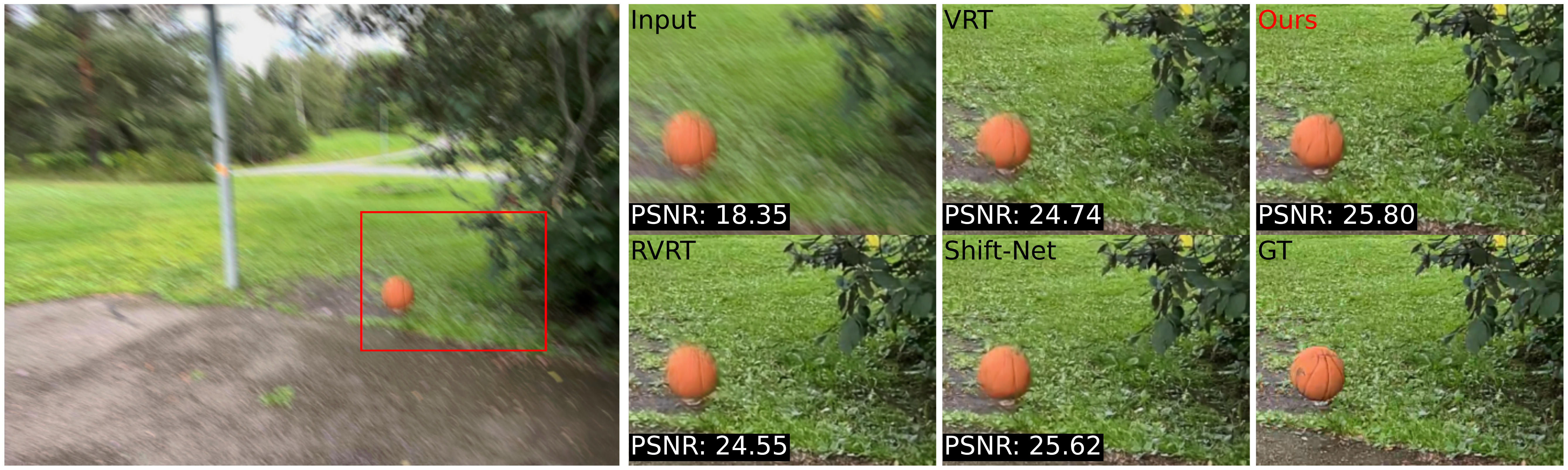}
    \end{center}
    \caption{Video deblurring examples from the SotA comparison.}
   \label{fig:SOTA_comparison}
\end{figure}

The results in \cref{tab:SOTA_comparison} validate our choice of using Shift-Net as the base model.
In single-image deblurring ($T=1$), our extended RGBD Shift-Net model not only surpasses the original Shift-Net in performance, as shown in \cref{sec:exp_depth_impact}, but also outperforms DGN in depth-aware deblurring requiring less computations.
Notably, the base structure of Shift-Net demonstrates greater robustness than DGN even without depth cues.
In video deblurring,
the performance of VRT is almost on par with the two Shift-Net variants, but their model is substantially larger and slower to train. 
Qualitative examples of video deblurring are shown in \cref{fig:SOTA_comparison}.
It is observed that RVRT provides a more natural curvature of the moving ball. However, our RGBD Shift-Net reveals finer textures within and sharper details around the ball, although the differences are minor.
\begin{table}[t]
    \centering
    \caption{SotA deblurring comparison using the \SHORTNAME test set. Single-image methods with $T=1$ and video-based methods with $T>1$. 'Depth' column indicates whether depth is used in the model.} 
    \resizebox{0.88\linewidth}{!}{
    \begin{tabular}{lccccrrr}
    \toprule
     & Depth & $T$ (frames) & PSNR & SSIM & GFLOPs & MParams & Training time [hrs] \\
    \midrule
    DGN~\cite{Li2020} & \checkmark & 1 & 23.94 & 0.8346 & 223.5 & 10.69 & 50 \\
    Base Shift-Net~\cite{Li2023} & \xmark & 1 & 25.28 & 0.8696 & 116.4 & 3.05 & 20 \\
    Our Shift-Net & \checkmark & 1 & 25.55 & 0.8735 & 177.7 & 4.17 & 30 \\
    \hline
    DGN~\cite{Li2020} & \checkmark & 5 & 25.63 & 0.8675 & 228.5 & 10.73 & 110 \\
    EDVR~\cite{Wang2019} & \xmark & 5 & 27.67 & 0.9098 & 778.1 & 23.60 & 90 \\
    RVRT~\cite{Liang2022b} & \xmark & 16 & 28.62 & 0.9286 & 933.9 & 13.57 & 370 \\
    VRT~\cite{Liang2022a} & \xmark & 6 & 29.03 & 0.9341 & 1118.4 & 18.03 & 290 \\
    Base Shift-Net~\cite{Li2023} & \xmark & 11 & 29.29 & 0.9353 & 313.2 & 3.05 & 110 \\
    Our Shift-Net & \checkmark & 11 & 29.33 & 0.9353 & 481.8 & 4.17 & 150 \\
    \bottomrule
    \end{tabular}}\\
    \label{tab:SOTA_comparison}
\end{table}
\subsection{Ablation studies}
\label{sec:ablation}
For faster training times the ablation studies were made using a smaller version of Shift-Net
(stack sizes $N_{s_1} = 1$, $N_{s_2} = 2$, $N_{s_3} = 1$, and $T=3$). 
Ablation performances are reported for \SHORTNAME \textit{validation data} to ensure that the test set results in  \cref{sec:exp_depth_impact} and \cref{sec:exp_sota} are unbiased.
\paragraph{Depth fusion blocks.}
We compared the effectiveness of our depth fusion method in \cref{sec:depth_injection} against competitive approaches.
As a baseline, we considered a simple fusion block where the depth and RGB features are simply concatenated and then processed by a convolution layer. This model is referred to as 'Concat+Conv'.
In addition, we included the original SFT and DAM blocks in DGN~\cite{Li2020} and DAST-Net\cite{Zhu2022a} since our DaT is partially inspired by those.
Each model was trained three times, and the mean and standard deviation results are summarized in \cref{tab:arch_ablation}.
The results show slightly better performance for DaT and GSS improves all except DAM.
\begin{table}[h]
  \caption{Comparison of depth fusion blocks in \cref{sec:depth_injection} with three input frames.}
  \label{tab:arch_ablation}
  \centering
  \resizebox{0.65\linewidth}{!}{%
  \begin{tabular}{lccc}
    \toprule
    Fusion Module & GSS & PSNR & SSIM \\
    \midrule
    Concat+Conv & \xmark & 24.37 ± 0.18 & 0.8078 ± 0.0051 \\
    Concat+Conv & \checkmark & 24.85 ± 0.11 & 0.8263 ± 0.0045 \\
    SFT & \xmark & 24.92 ± 0.12 & 0.8277 ± 0.0051 \\
    SFT & \checkmark & 25.10 ± 0.08 & 0.8351 ± 0.0030 \\
    DAM & \xmark & 25.12 ± 0.06 & 0.8360 ± 0.0015 \\
    DAM & \checkmark & 25.07 ± 0.01 & 0.8344 ± 0.0008 \\
    DaT & \xmark & 25.12 ± 0.12 & 0.8376 ± 0.0038 \\
    DaT & \checkmark & \textbf{25.16 ± 0.06} & \textbf{0.8381 ± 0.0016} \\
    \bottomrule
  \end{tabular}}
\end{table}
\paragraph{Depth quality.}
Another important aspect is whether the noisy measurements from a depth sensor provide advantage over monocular depth. %
Two variants of monocular depth maps were generated with the Kim~\etal~\cite{kim2022globallocal} method.
Monocular depth estimated from blurry inputs represents a more realistic case while monocular depth from the sharp RGB represents an ideal case as it uses the ground truth frames. 
Kim~\etal was used since its code is publicly available and obtains high performance on multiple datasets. 
In particular, we used pre-trained weights from the NYUv2 dataset~\cite{Silberman2012}.

The results in \cref{tab:depth_quality} verify that including real depth measurements is beneficial for deblurring. Moreover, the superior results with sharp monocular depth ('ideal case') indicate that sharp details might be more critical than depth \textit{per se}.
These findings are illustrated in \cref{fig:depth_quality}.
\begin{table}[t]
  \caption{Evaluation of depth-aware deblurring with LiDAR sensor (iPhone) and monocular depth estimation.}
  \label{tab:depth_quality}
  \centering
 \resizebox{0.7\linewidth}{!}{%
  \begin{tabular}{lccc}
    \toprule
    Input & Monocular (blur) & LiDAR (iPhone) & Monocular (sharp)$^\dagger$ \\
    \midrule
    PSNR & 25.22 & 25.24 & 25.37 \\
    SSIM & 0.840 & 0.842 & 0.846 \\
    \bottomrule
  \end{tabular}}\\
  {\small $^\dagger$ ideal case as the sharp inputs are also groundtruth}
\end{table}
%
%
\section{Discussion and Limitations}
The application of the methods introduced in this paper are limited to devices where depth is available. 
However, this research work intends to unveil the benefits of incorporating depth information, beyond the specific depth-aware deblurring architecture that is proposed.
Consequently, we offer insights that could assist camera manufacturers in determining the value of integrating depth sensors into their image processing pipelines. Given the increasing availability of RGB-D devices, our findings are relevant and timely.

\begin{figure}[t]
    \begin{center}\includegraphics[width=0.8\linewidth]{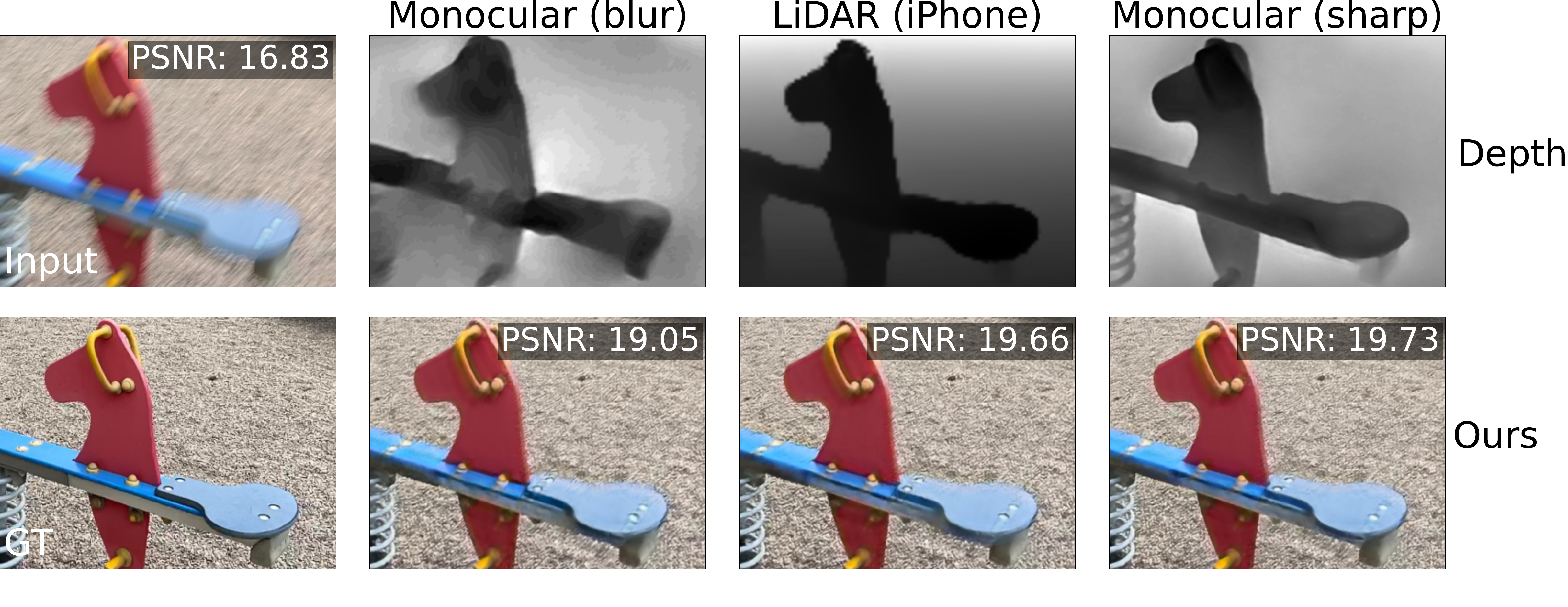}
    \end{center}
    \caption{Video deblurring examples using different depth inputs.}
   \label{fig:depth_quality}
\end{figure}

Our results in \cref{sec:exp_depth_impact} demonstrated that unreliable depth measurements do not negatively impact our RGBD Shift-Net model compared to a baseline model that does not use depth inputs. 
Nevertheless, these results may not generalize to other depth sensors, as the depth maps from the iPhone's ARKit, enhanced with monocular depth, might be more robust against highly reflective and absorbent surfaces.
Additionally, the depth resolution in the DAVIDE dataset is relatively low compared to the image resolution and those offered by other sensors in the market. 
For instance, \cref{fig:depth_quality} illustrates the lack of sharp details in our sensed depth compared to the depth map obtained from pure monocular depth estimation with the sharp RGB image.
Despite this lower resolution, our RGBD Shift-Net model still benefits from the available depth information.

We showed that combining our DaT block with the GSS block results in the most effective model for depth-aware video deblurring in \cref{sec:ablation}.
Notwithstanding, such depth injection methods increase the computational complexity by 53.8\% (for $T=11$ frames) and the number of parameters by 36.7\% compared to the RGB-only architecture (refer to \cref{tab:SOTA_comparison}).
Although this additional complexity may restrict their use in real-time or resource-limited environments, it remains significantly lower than other SotA methods like VRT or RVRT.
%
%
\section{Conclusion}
We proposed a new dataset (\SHORTNAME{}) and a robust network architecture for depth-aware video deblurring. 
Our architecture utilizes a depth injection method including the newly proposed DaT block, to incorporate depth information into the existing RGB video deblurring method Shift-Net.
With their help, we provided novel insights into the use of depth in video processing. 
Our findings indicate that video deblurring methods can compensate for the lack of depth information by accessing longer context windows.
In the case of single-image deblurring, indoor, close proximity, and camera motion scenes clearly benefit from depth. 
The advantage in indoor and close proximity conditions is due to the LiDAR sensor's higher accuracy in these conditions.
For scenarios with only camera motion, we believe depth cues could be more easily used to resolve parallax blur.
Finally, depth maps might not inform the amount of motion blur, but instead, they supply edges and structures that help deblurring methods reconstruct sharp details.

\vspace{\medskipamount}
\noindent\textbf{Acknowledgements.} 
This project was supported by a Huawei Technologies Oy (Finland) project. The authors thank Jussi Yliayho for general project discussions and Lauri Suomela for his valuable comments on a previous version of the manuscript.
%
%
%
%
\bibliographystyle{splncs04}
\bibliography{main}

\begin{thebibliography}{10}
\providecommand{\url}[1]{\texttt{#1}}
\providecommand{\urlprefix}{URL }
\providecommand{\doi}[1]{https://doi.org/#1}

\bibitem{Bar2007}
Bar, L., Berkels, B., Rumpf, M., Sapiro, G.: A variational framework for
  simultaneous motion estimation and restoration of motion-blurred video. In:
  2007 IEEE 11th International Conference on Computer Vision. pp.~1--8. IEEE
  (2007)

\bibitem{Cao2022}
Cao, M., Fan, Y., Zhang, Y., Wang, J., Yang, Y.: Vdtr: Video deblurring with
  transformer. IEEE Transactions on Circuits and Systems for Video Technology
  \textbf{33}(1),  160--171 (2022)

\bibitem{Chambolle2011}
Chambolle, A., Pock, T.: A first-order primal-dual algorithm for convex
  problems with applications to imaging. Journal of mathematical imaging and
  vision  \textbf{40},  120--145 (2011)

\bibitem{Chen2022}
Chen, L., Chu, X., Zhang, X., Sun, J.: Simple baselines for image restoration.
  In: European conference on computer vision. pp. 17--33. Springer (2022)

\bibitem{Cho2012}
Cho, S., Wang, J., Lee, S.: Video deblurring for hand-held cameras using
  patch-based synthesis. ACM Transactions on Graphics (TOG)  \textbf{31}(4),
  ~1--9 (2012)

\bibitem{Chu2022}
Chu, X., Chen, L., Chen, C., Lu, X.: Improving image restoration by revisiting
  global information aggregation. In: European Conference on Computer Vision.
  pp. 53--71. Springer (2022)

\bibitem{Cordts2016}
Cordts, M., Omran, M., Ramos, S., Rehfeld, T., Enzweiler, M., Benenson, R.,
  Franke, U., Roth, S., Schiele, B.: The cityscapes dataset for semantic urban
  scene understanding. In: Proc. of the IEEE Conference on Computer Vision and
  Pattern Recognition (CVPR) (2016)

\bibitem{Debevec-CRF}
Debevec, P.E., Malik, J.: Recovering high dynamic range radiance maps from
  photographs. In: Proceedings of the 24th Annual Conference on Computer
  Graphics and Interactive Techniques. p. 369–378. SIGGRAPH '97, ACM
  Press/Addison-Wesley Publishing Co., USA (1997). \doi{10.1145/258734.258884},
  \url{https://doi.org/10.1145/258734.258884}

\bibitem{Feng2023}
Feng, Y., Hansen, P., Whatmough, P.N., Lu, G., Zhu, Y.: Fast and accurate:
  Video enhancement using sparse depth. In: Proceedings of the IEEE/CVF Winter
  Conference on Applications of Computer Vision. pp. 4492--4500 (2023)

\bibitem{Geiger2013}
Geiger, A., Lenz, P., Stiller, C., Urtasun, R.: Vision meets robotics: The
  kitti dataset. International Journal of Robotics Research (IJRR)  (2013)

\bibitem{Hu2014}
Hu, Z., Xu, L., Yang, M.H.: Joint depth estimation and camera shake removal
  from single blurry image. In: Proceedings of the IEEE Conference on Computer
  Vision and Pattern Recognition. pp. 2893--2900 (2014)

\bibitem{Hyun2015}
Hyun~Kim, T., Mu~Lee, K.: Generalized video deblurring for dynamic scenes. In:
  Proceedings of the IEEE Conference on Computer Vision and Pattern
  Recognition. pp. 5426--5434 (2015)

\bibitem{Hyun2017}
Hyun~Kim, T., Mu~Lee, K., Scholkopf, B., Hirsch, M.: Online video deblurring
  via dynamic temporal blending network. In: Proceedings of the IEEE
  international conference on computer vision. pp. 4038--4047 (2017)

\bibitem{Jiang2022}
Jiang, B., Xie, Z., Xia, Z., Li, S., Liu, S.: Erdn: Equivalent receptive field
  deformable network for video deblurring. In: European Conference on Computer
  Vision. pp. 663--678. Springer (2022)

\bibitem{SSloMo}
Jiang, H., Sun, D., Jampani, V., Yang, M., Learned{-}Miller, E.G., Kautz, J.:
  Super slomo: High quality estimation of multiple intermediate frames for
  video interpolation. In: CVPR (2018)

\bibitem{kim2022globallocal}
Kim, D., Ka, W., Ahn, P., Joo, D., Chun, S., Kim, J.: Global-local path
  networks for monocular depth estimation with vertical cutdepth (2022)

\bibitem{Kim2018}
Kim, T.H., Sajjadi, M.S., Hirsch, M., Scholkopf, B.: Spatio-temporal
  transformer network for video restoration. In: Proceedings of the European
  conference on computer vision (ECCV). pp. 106--122 (2018)

\bibitem{Krishnan2009}
Krishnan, D., Fergus, R.: Fast image deconvolution using hyper-laplacian
  priors. Advances in neural information processing systems  \textbf{22} (2009)

\bibitem{Li2023}
Li, D., Shi, X., Zhang, Y., Cheung, K.C., See, S., Wang, X., Qin, H., Li, H.: A
  simple baseline for video restoration with grouped spatial-temporal shift.
  In: Proceedings of the IEEE/CVF Conference on Computer Vision and Pattern
  Recognition. pp. 9822--9832 (2023)

\bibitem{Li2021}
Li, D., Xu, C., Zhang, K., Yu, X., Zhong, Y., Ren, W., Suominen, H., Li, H.:
  Arvo: Learning all-range volumetric correspondence for video deblurring. In:
  Proceedings of the IEEE/CVF Conference on Computer Vision and Pattern
  Recognition. pp. 7721--7731 (2021)

\bibitem{Li2020}
Li, L., Pan, J., Lai, W.S., Gao, C., Sang, N., Yang, M.H.: Dynamic scene
  deblurring by depth guided model. IEEE Transactions on Image Processing
  \textbf{29},  5273--5288 (2020). \doi{10.1109/TIP.2020.2980173}

\bibitem{Liang2022a}
Liang, J., Cao, J., Fan, Y., Zhang, K., Ranjan, R., Li, Y., Timofte, R.,
  Van~Gool, L.: Vrt: A video restoration transformer. arXiv preprint
  arXiv:2201.12288  (2022)

\bibitem{Liang2022b}
Liang, J., Fan, Y., Xiang, X., Ranjan, R., Ilg, E., Green, S., Cao, J., Zhang,
  K., Timofte, R., Gool, L.V.: Recurrent video restoration transformer with
  guided deformable attention. Advances in Neural Information Processing
  Systems  \textbf{35},  378--393 (2022)

\bibitem{Lin2022}
Lin, J., Cai, Y., Hu, X., Wang, H., Yan, Y., Zou, X., Ding, H., Zhang, Y.,
  Timofte, R., Van~Gool, L.: Flow-guided sparse transformer for video
  deblurring. arXiv preprint arXiv:2201.01893  (2022)

\bibitem{SIFT-paper}
Lowe, D.G.: Distinctive image features from scale-invariant keypoints.
  International journal of computer vision  \textbf{60},  91--110 (2004)

\bibitem{Nah2019a}
Nah, S., Baik, S., Hong, S., Moon, G., Son, S., Timofte, R., Mu~Lee, K.: Ntire
  2019 challenge on video deblurring and super-resolution: Dataset and study.
  In: Proceedings of the IEEE/CVF Conference on Computer Vision and Pattern
  Recognition Workshops. pp.~0--0 (2019)

\bibitem{Nah2017}
Nah, S., Hyun~Kim, T., Mu~Lee, K.: Deep multi-scale convolutional neural
  network for dynamic scene deblurring. In: Proceedings of the IEEE conference
  on computer vision and pattern recognition. pp. 3883--3891 (2017)

\bibitem{Nah2019b}
Nah, S., Son, S., Lee, K.M.: Recurrent neural networks with intra-frame
  iterations for video deblurring. In: Proceedings of the IEEE/CVF conference
  on computer vision and pattern recognition. pp. 8102--8111 (2019)

\bibitem{Niklaus-2017-iccv}
Niklaus, S., Mai, L., Liu, F.: Video frame interpolation via adaptive separable
  convolution. In: ICCV (2017)

\bibitem{Pan2020}
Pan, J., Bai, H., Tang, J.: Cascaded deep video deblurring using temporal
  sharpness prior. In: Proceedings of the IEEE/CVF conference on computer
  vision and pattern recognition. pp. 3043--3051 (2020)

\bibitem{Pan2023}
Pan, J., Xu, B., Dong, J., Ge, J., Tang, J.: Deep discriminative spatial and
  temporal network for efficient video deblurring. In: Proceedings of the
  IEEE/CVF Conference on Computer Vision and Pattern Recognition. pp.
  22191--22200 (2023)

\bibitem{Pan2019}
Pan, L., Dai, Y., Liu, M.: Single image deblurring and camera motion estimation
  with depth map. Proceedings - 2019 IEEE Winter Conference on Applications of
  Computer Vision, WACV 2019 pp. 2116--2125 (3 2019).
  \doi{10.1109/WACV.2019.00229}

\bibitem{Park2017}
Park, H., Mu~Lee, K.: Joint estimation of camera pose, depth, deblurring, and
  super-resolution from a blurred image sequence. In: Proceedings of the IEEE
  International Conference on Computer Vision. pp. 4613--4621 (2017)

\bibitem{Rim2020}
Rim, J., Lee, H., Won, J., Cho, S.: Real-world blur dataset for learning and
  benchmarking deblurring algorithms. In: Computer Vision--ECCV 2020: 16th
  European Conference, Glasgow, UK, August 23--28, 2020, Proceedings, Part XXV
  16. pp. 184--201. Springer (2020)

\bibitem{Robertson-CRF}
Robertson, M.A., Borman, S., Stevenson, R.L.: Dynamic range improvement through
  multiple exposures. In: ICIP (1999)

\bibitem{Scales1988}
Scales, J.A., Gersztenkorn, A.: Robust methods in inverse theory. Inverse
  Problems  \textbf{4}(4), ~1071 (oct 1988). \doi{10.1088/0266-5611/4/4/010},
  \url{https://dx.doi.org/10.1088/0266-5611/4/4/010}

\bibitem{Shen2019}
Shen, Z., Wang, W., Lu, X., Shen, J., Ling, H., Xu, T., Shao, L.: Human-aware
  motion deblurring. In: Proceedings of the IEEE/CVF International Conference
  on Computer Vision. pp. 5572--5581 (2019)

\bibitem{Sheng2019}
Sheng, B., Li, P., Fang, X., Tan, P., Wu, E.: Depth-aware motion deblurring
  using loopy belief propagation. IEEE Transactions on Circuits and Systems for
  Video Technology  \textbf{30}(4),  955--969 (2019)

\bibitem{Silberman2012}
Silberman, N., Hoiem, D., Kohli, P., Fergus, R.: Indoor segmentation and
  support inference from rgbd images. In: Computer Vision--ECCV 2012: 12th
  European Conference on Computer Vision, Florence, Italy, October 7-13, 2012,
  Proceedings, Part V 12. pp. 746--760. Springer (2012)

\bibitem{Sim2021}
Sim, H., Oh, J., Kim, M.: Xvfi: extreme video frame interpolation. In:
  Proceedings of the IEEE/CVF international conference on computer vision. pp.
  14489--14498 (2021)

\bibitem{Son2021}
Son, H., Lee, J., Lee, J., Cho, S., Lee, S.: Recurrent video deblurring with
  blur-invariant motion estimation and pixel volumes. ACM Transactions on
  Graphics (TOG)  \textbf{40}(5),  1--18 (2021)

\bibitem{Sturm2012}
Sturm, J., Engelhard, N., Endres, F., Burgard, W., Cremers, D.: A benchmark for
  the evaluation of rgb-d slam systems. In: Proc. of the International
  Conference on Intelligent Robot Systems (IROS) (Oct 2012)

\bibitem{Su2017}
Su, S., Delbracio, M., Wang, J., Sapiro, G., Heidrich, W., Wang, O.: Deep video
  deblurring for hand-held cameras. In: Proceedings of the IEEE conference on
  computer vision and pattern recognition. pp. 1279--1288 (2017)

\bibitem{Suin2021}
Suin, M., Rajagopalan, A.: Gated spatio-temporal attention-guided video
  deblurring. In: Proceedings of the IEEE/CVF Conference on Computer Vision and
  Pattern Recognition. pp. 7802--7811 (2021)

\bibitem{Torres2023}
Torres, G.F., K{\"a}m{\"a}r{\"a}inen, J.: Depth-aware image compositing model
  for parallax camera motion blur. In: Scandinavian Conference on Image
  Analysis. pp. 279--296. Springer (2023)

\bibitem{Wang2021}
Wang, T., Zhang, X., Jiang, R., Zhao, L., Chen, H., Luo, W.: Video deblurring
  via spatiotemporal pyramid network and adversarial gradient prior. Computer
  Vision and Image Understanding  \textbf{203},  103135 (2021)

\bibitem{Wang2019}
Wang, X., Chan, K.C., Yu, K., Dong, C., Change~Loy, C.: Edvr: Video restoration
  with enhanced deformable convolutional networks. In: Proceedings of the
  IEEE/CVF conference on computer vision and pattern recognition workshops.
  pp.~0--0 (2019)

\bibitem{Wang2018}
Wang, X., Yu, K., Dong, C., Loy, C.C.: Recovering realistic texture in image
  super-resolution by deep spatial feature transform. In: Proceedings of the
  IEEE conference on computer vision and pattern recognition. pp. 606--615
  (2018)

\bibitem{Wulff2014}
Wulff, J., Black, M.J.: Modeling blurred video with layers. In: Computer
  Vision--ECCV 2014: 13th European Conference, Zurich, Switzerland, September
  6-12, 2014, Proceedings, Part VI 13. pp. 236--252. Springer (2014)

\bibitem{Xiang2020}
Xiang, X., Wei, H., Pan, J.: Deep video deblurring using sharpness features
  from exemplars. IEEE Transactions on Image Processing  \textbf{29},
  8976--8987 (2020)

\bibitem{Xu2012}
Xu, L., Jia, J.: Depth-aware motion deblurring. 2012 IEEE International
  Conference on Computational Photography, ICCP 2012  (2012).
  \doi{10.1109/ICCPHOT.2012.6215220}

\bibitem{Xu2021}
Xu, Q., Pan, J., Qian, Y.: Learning an occlusion-aware network for video
  deblurring. IEEE Transactions on Circuits and Systems for Video Technology
  \textbf{32}(7),  4312--4323 (2021)

\bibitem{Zamir2022}
Zamir, S.W., Arora, A., Khan, S., Hayat, M., Khan, F.S., Yang, M.H.: Restormer:
  Efficient transformer for high-resolution image restoration. In: Proceedings
  of the IEEE/CVF conference on computer vision and pattern recognition. pp.
  5728--5739 (2022)

\bibitem{EMA-VFI}
Zhang, G., Zhu, Y., Wang, H., Chen, Y., Wu, G., Wang, L.: Extracting motion and
  appearance via inter-frame attention for efficient video frame interpolation
  (2023)

\bibitem{Zhang2022}
Zhang, H., Xie, H., Yao, H.: Spatio-temporal deformable attention network for
  video deblurring. In: European Conference on Computer Vision. pp. 581--596.
  Springer (2022)

\bibitem{Zhang2018}
Zhang, K., Luo, W., Zhong, Y., Ma, L., Liu, W., Li, H.: Adversarial
  spatio-temporal learning for video deblurring. IEEE Transactions on Image
  Processing  \textbf{28}(1),  291--301 (2018)

\bibitem{Zhang2016}
Zhang, L., Zhou, L., Huang, H.: Bundled kernels for nonuniform blind video
  deblurring. IEEE Transactions on Circuits and Systems for Video Technology
  \textbf{27}(9),  1882--1894 (2016)

\bibitem{Zhang2020}
Zhang, X., Jiang, R., Wang, T., Wang, J.: Recursive neural network for video
  deblurring. IEEE Transactions on Circuits and Systems for Video Technology
  \textbf{31}(8),  3025--3036 (2020)

\bibitem{Zhang2021}
Zhang, X., Wang, T., Jiang, R., Zhao, L., Xu, Y.: Multi-attention convolutional
  neural network for video deblurring. IEEE Transactions on Circuits and
  Systems for Video Technology  \textbf{32}(4),  1986--1997 (2021)

\bibitem{Zhong2020}
Zhong, Z., Gao, Y., Zheng, Y., Zheng, B.: Efficient spatio-temporal recurrent
  neural network for video deblurring. In: Computer Vision--ECCV 2020: 16th
  European Conference, Glasgow, UK, August 23--28, 2020, Proceedings, Part VI
  16. pp. 191--207. Springer (2020)

\bibitem{Zhong2023}
Zhong, Z., Gao, Y., Zheng, Y., Zheng, B., Sato, I.: Real-world video
  deblurring: A benchmark dataset and an efficient recurrent neural network.
  International Journal of Computer Vision  \textbf{131}(1),  284--301 (2023)

\bibitem{Zhou2019b}
Zhou, S., Zhang, J., Pan, J., Xie, H., Zuo, W., Ren, J.: Spatio-temporal filter
  adaptive network for video deblurring. In: Proceedings of the IEEE/CVF
  international conference on computer vision. pp. 2482--2491 (2019)

\bibitem{Zhu2022b}
Zhu, C., Dong, H., Pan, J., Liang, B., Huang, Y., Fu, L., Wang, F.: Deep
  recurrent neural network with multi-scale bi-directional propagation for
  video deblurring. In: Proceedings of the AAAI conference on artificial
  intelligence. pp. 3598--3607 (2022)

\bibitem{Zhu2022a}
Zhu, Q., Xiao, Z., Huang, J., Zhao, F.: Dast-net: Depth-aware spatio-temporal
  network for video deblurring. Proceedings - IEEE International Conference on
  Multimedia and Expo  \textbf{2022-July} (2022).
  \doi{10.1109/ICME52920.2022.9858929}

\end{thebibliography}
%
%
%
\clearpage
\appendix
\appendixname
\pagenumbering{arabic}
\renewcommand*{\thepage}{A\arabic{page}}
\counterwithin{figure}{subsection}
\counterwithin{table}{subsection}
\counterwithin{equation}{subsection}
%
%
\section{DAVIDE dataset: Further details}
\subsection{CRF calibration}
\label{sec:crf_calibration}
\begin{wrapfigure}{r}{0.4\textwidth}
  \centering
  \includegraphics[width=\linewidth]{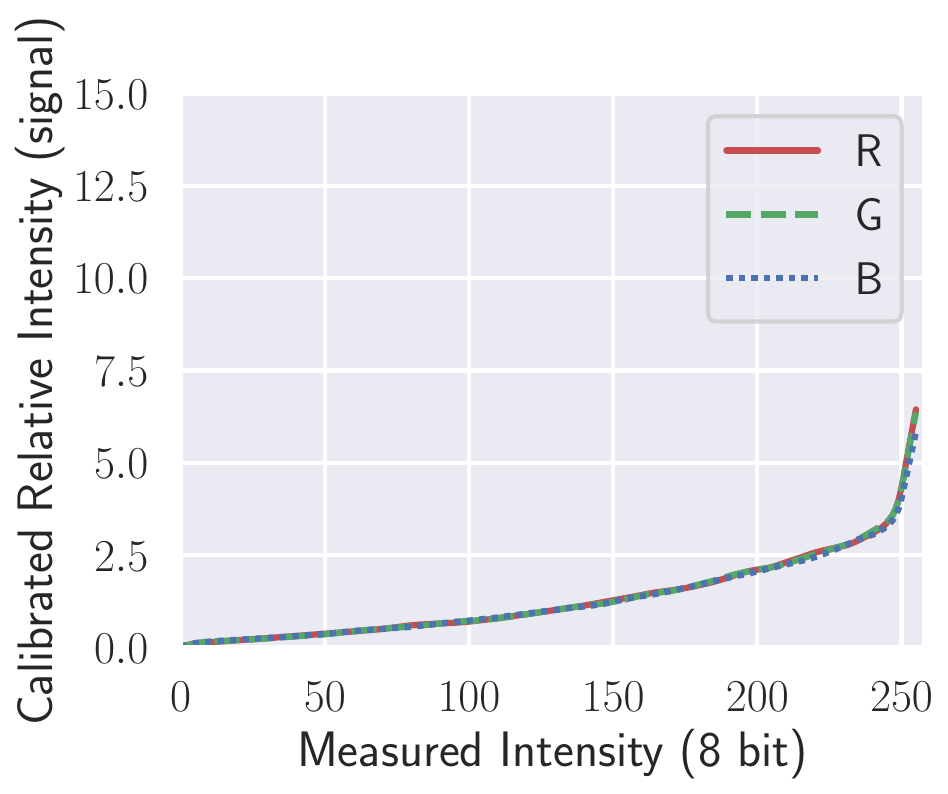}
  \caption{Inverse CRF used in our DAVIDE synthesis pipeline.}
  \label{fig:crf_davide}
\end{wrapfigure}

We calibrated the inverse camera response function (CRF) of our capture device using the Debevec algorithm~\cite{Debevec-CRF}.
This process requires capturing images of a static scene at various exposure times, \aka exposure bracketing.
To ensure accurate CRF calibration, we secured the capturing device on a tripod and performed computational image alignment post-capture, before running the calibration algorithm. 
Specifically, we warped the images towards a reference frame using estimated homography transformations computed from matched SIFT feature points~\cite{SIFT-paper}.
The Debevec algorithm was then applied to these aligned images. 
To cope with the inaccuracies in the saturated region, we made a linear approximation following to the calibration procedure of the REDS dataset~\cite{Nah2019a}:
\begin{equation}
    \label{eq:linear_approx_CRF}
    \textit{CRF}^{-1}(p) = \begin{cases}
        \Gamma(p) & p \le 250\\
        m (p-250) + \Gamma(250) & p > 250
    \end{cases}
\end{equation}
where $\Gamma$ denotes the inverse CRF obtained through the Debevec algorithm, $p$ is the pixel value of the color channel, and $m$ is the linear approximation calculated as $m = \tfrac{\Gamma(251) - \Gamma(249)}{2}$.
\cref{fig:crf_davide} depicts the inverse CRF that is obtained and used in the synthesis pipeline.
%
%
\section{Shift-Net and RGBD extension}

\subsection{Depth-aware Transformer Block (DaT)}
\label{sec:dat}
\cref{fig:depth_fusion} in the main manuscript illustrates the Depth-aware Transformer Block (DaT) that is used for depth fusion in our depth-extended Shift-Net architecture.
Here, we provide a more detailed description of the DaT block.
The DaT block consists of three main components: 1) a cross-attention module to adapt the depth features based on reference RGB features, 2) a Spatial Feature Transform (SFT) layer to module the RGB features based on the attention depth features, and 3) a gated feed-forward network to control the feature aggregation. The DaT block is inspired by the Restormer architecture~\cite{Zamir2022}, which incorporates the multi-Dconv head transposed attention (MDTA) module and a gated-Dconv feed-forward network (GDFN).
While our gated feed-forward network corresponds to the same GDFN block as in Restormer, the cross-attention module is an adaptation of the MDTA module to perform cross-attention with the depth features.

\paragraph{Cross-attention module.}
Given a reference RGB feature frame $F_I[m] \in \mathbb{R}^{\hat{H} \times \hat{W} \times C_I}$ and a supporting depth feature frame $F'_D[m] \in \mathbb{R}^{\hat{H} \times \hat{W} \times C_D}$, where $\hat{H} \times \hat{W}$ is the spatial resolution of the level, and $C_I$,$C_D$ denote the RGB and depth feature dimensions, respectively;
we compute \textit{query} $Q$, \textit{key} $K$, and \textit{value} $V$ features from $F_I$ and $F'_D$ as:
\begin{align}
    Q &= W_Q \cdot F_I, \quad K = W_K \cdot F'_D, \quad V = W_V \cdot F'_D
\end{align}
where $W_Q$, $W_K$, and $W_V$ are bias-free convolutional layers comprised of a 1x1 convolution followed by a 3x3 depth-wise convolution. Then, we reshape the \textit{query} and \textit{key} features such that their dot product yields to an attention map of size $C_I \times C_D$, whereas the \textit{value} features are reshaped so that its reweighted sum is performed over the channel dimension.
Accordingly, the cross-attention module produces modulated depth features as follows:
\begin{align}
  \label{eq:cross_attention}
  \begin{split}
      Q &\rightarrow \mathbb{R}^{\hat{H} \hat{W} \times C_I}, \quad K \rightarrow \mathbb{R}^{\hat{H} \hat{W} \times C_D}, \quad V \rightarrow \mathbb{R}^{\hat{H} \hat{W} \times C_D} \quad \text{\texttt{\# reshape tensors}} \\
      A &= \texttt{softmax}(\tfrac{Q^T K}{\alpha}), \quad A \in \mathbb{R}^{C_I \times C_D} \quad \text{\texttt{\# compute attention map}} \\
      Y &= V A^T, \quad Y \in \mathbb{R}^{\hat{H} \hat{W} \times C_I} \quad \text{\texttt{\# reweighted sum}}\\
      Y &\rightarrow \mathbb{R}^{\hat{H} \times \hat{W} \times C_I} \quad \text{\texttt{\# reshape tensor}}\\
      F_{D \rightarrow I} &= W_p \cdot Y \quad \text{\texttt{\# final projection}}
  \end{split}
  \end{align}
  where $F_{D \rightarrow I}$ denotes the aligned features, $W_p$ is a 1x1 convolution layer for final aggregation, and $\alpha$ is a learnable scaling parameter that controls the magnitude of the dot product.
  Since $Q$ and $K$ come from $F_I$ and $F'_D$, respectively, $A$ contains the correlation coefficients between the channels of the reference RGB features and the channels in the supporting depth features.
  Therefore, this module yields to depth features that are aligned with the RGB features in the query.
  Following the multi-head strategy, the number of channels is divided into 'heads' to learn several attention maps in parallel.

\paragraph{Spatial Feature Transform (SFT).}
We use the SFT layer in our architecture to modulate the RGB features $F_I$ based on a condition signal $z$. In our case, the aligned depth features $F_{D \rightarrow I}$ produced by the cross-attention module are used as the condition signal:
\begin{align}
  \label{eq:SFT}
  \begin{split}
      \gamma &= \mathcal{M}_\gamma (z)|_{z=F_{D \rightarrow I}} \quad \text{\texttt{\# scaling tensor}}\\
      \beta &= \mathcal{M}_\beta (z)|_{z=F_{D \rightarrow I}} \quad \text{\texttt{\# offset tensor}}\\
      \ddot{F}_I &= \gamma \odot F_I + \beta \quad \text{\texttt{\# modulation}}
  \end{split}
  \end{align}
where $\odot$ denotes the element-wise multiplication, and $\mathcal{M}_\gamma$, $\mathcal{M}_\beta$ are mapping functions obtained through a small convolutional block for the scaling $\gamma$ and offset $\beta$, respectively.

\paragraph{Gated feed-forward network.}
Given a modulated RGB feature $\ddot{F}_I$, this module further controls the feature aggregation as:
\begin{equation}
    \tilde{F}_I = W^0 (\sigma(W^1(\text{LN}(\ddot{F}_I)) \odot W^2(\text{LN}(\ddot{F}_I)))) + \ddot{F}_I
\end{equation}
where $\sigma$ denotes the GELU non-linearity, $\text{LN}$ is a normalization layer, and $W^{(\cdot)}$ represent convolutional layers.

\subsection{Training and inference details}
\paragraph{Training.}
The original Shift-Net architecture and our RGBD extension are in nature video deblurring methods that leverage temporal correlation among consecutive frames to recover the sharp details. 
In our experiments, we trained both models with a varying temporal length $T$ of the context window, including $T=1$ for single-image deblurring. 
To handle the single-image case, we needed to slightly adjust the architectures as the video-based versions take $T$ input frames to produce $T-2$ frames in the output.
Our solution was to omit the temporal shift in \cref{eq:temporal_shift} while maintaining the rest of the architecture unchanged, as the remaining blocks can process frames individually.

Regarding the training recipe, our dataloader uses a customized sampler that randomly selects 100 sequences from each video clip in the training set. 
Depth maps provided by the sensor are normalized by dividing each by the maximum depth value in the sequence. 
We used mean absolute error (MAE) for the loss function and enhanced training efficiency with automatic mixed precision and gradient clipping set to a value of 1.

\paragraph{Inference.}
Video-based RGB-only and RGBD Shift-Net models can process a long sequence of frames at a time, regardless of the number of frames that are used during training. 
This flexibility is due to their reuse of the same frame-wise processing components and reliance on a series of forward and backward shifts for temporal aggregation.
Therefore, we take 27 input frames and generate 25 output frames at a time during inference.
We also employed patch-based processing with patches of 640 pixels and an overlap of 16 pixels to manage large inputs efficiently. 
To reduce the distribution shift of training and inference, we implemented the local average pooling of the Test-time Local Converter (TLC)~\cite{Chu2022}.
Likewise, we enhanced the inference efficiency with automatic mixed precision.
%
%
\section{State-of-the-Art comparison}
\subsection{Training settings}
\label{sec:sota_training_settings}
\begin{table}[t]
\centering
\caption{Training settings used in our SotA comparison.}
\resizebox{0.98\linewidth}{!}{
\begin{tabular}{llllll}
\toprule
                  & DGN~\cite{Li2020}       & EDVR~\cite{Wang2019}      & RVRT~\cite{Liang2022b}      & VRT~\cite{Liang2022a}       & Shift-Net~\cite{Li2023}  \\
\midrule
patch size        & 304       & 256       & 256       & 192       & 256        \\
epochs            & 300       & 600       & 300       & 150       & 200        \\
warm-up           & 1         & -         & -         & -         & -          \\
batch size        & 4         & 16        & 4         & 4         & 4          \\
optimizer         & AdamW     & Adam      & Adam      & Adam      & AdamW      \\
lr                & $2 \times 10^{-4}$      & $4 \times 10^{-4}$      & $4 \times 10^{-5}$      & $1 \times 10^{-4}$      & $3 \times 10^{-4}$       \\
weight decay      & $1 \times 10^{-4}$      & 0.0       & 0.0       & 0.0       & 0.0        \\
betas             & 0.9, 0.99 & 0.9, 0.99 & 0.9, 0.99 & 0.9, 0.99 & 0.9, 0.99  \\
\multirow{3}{*}{scheduler}    & cosine annealing         & \begin{tabular}[t]{@{}l@{}}cosine annealing\\ with restart\end{tabular}    & cosine annealing & cosine annealing   & cosine annealing \\
   & & \begin{tabular}[t]{@{}l@{}}periods: \\ 25, 100, 150, 150, 175\end{tabular} &   &   &   \\
    &   & \begin{tabular}[t]{@{}l@{}}weights:\\ 1, 0.8, 0.5, 0.5, 0.2\end{tabular}   &    &    &   \\
min. lr           & $1 \times 10^{-7}$     & $1 \times 10^{-7}$      & $1 \times 10^{-7}$       & $1 \times 10^{-7}$      & $1 \times 10^{-7}$     \\
gradient clip     & 1.0      & -         & -          & -         & 1.0      \\
EMA$^\dagger$ decay         & 0.0      & 0.999     & 0.0        & 0.0       & 0.0      \\
enable AMP$^\mathsection$        & True     & False     & False      & False     & True     \\
pixel loss        & MAE   & Charbonnier loss & Charbonnier loss & Charbonnier loss & MAE \\
\multirow{2}{*}{perceptual loss} & vgg19   & -  & -  & -  & -  \\
                  & weight: $1 \times 10^{-4}$ &         &    &    &    \\
computing units   & \begin{tabular}[t]{@{}l@{}}2xNvidia V100\\ (32 GB)\end{tabular} & \begin{tabular}[t]{@{}l@{}}4xNvidia V100\\ (32 GB)\end{tabular}            & \begin{tabular}[t]{@{}l@{}}4xNvidia V100\\ (32 GB)\end{tabular} & \begin{tabular}[t]{@{}l@{}}4xNvidia A100\\ (40 GB)\end{tabular} & \begin{tabular}[t]{@{}l@{}}4xNvidia V100\\ (32 GB)\end{tabular}\\
\bottomrule
\end{tabular}}\\
{\scriptsize $^\dagger$ Exponential Moving Average (EMA)}\\
{\scriptsize $^\mathsection$ Automatic Mixed Precision (AMP)}\\
\label{tab:training_settings}
\end{table}

For reproducibility, \cref{tab:training_settings} details the training settings used in our state-of-the-art comparison experiment. Both our RGBD and RGB-only Shift-Net models used identical settings. In general, we adhered to the default settings from the original works but fine-tuned them to ensure robust convergence on the DAVIDE dataset.

\end{document}